\documentclass[runningheads]{llncs}
\usepackage{amsmath}
\usepackage{amssymb}
\usepackage{graphicx}
\usepackage{booktabs}
\usepackage{subfig}
\usepackage{tabularx}
\usepackage[colorlinks=true,urlcolor=blue,citecolor=blue,linkcolor=blue,pdfborder=0 0 0]{hyperref}

\usepackage[strings]{underscore}
\usepackage{butterma2021}

\begin{document}
\title{Multi-view Analysis of Unregistered Medical Images Using Cross-View Transformers}
\titlerunning{Multi-view Analysis Using Cross-View Transformers}
\author{%
  Gijs van Tulder \and 
  Yao Tong \and
  Elena Marchiori 
}
\hypersetup{%
  pdfinfo={%
    Title={Multi-view Analysis of Unregistered Medical Images Using Cross-View Transformers},
    Author={Gijs van Tulder, Yao Tong, Elena Marchiori}
  }%
}
\authorrunning{G. van Tulder et al.}
\institute{%
  Data Science Group, Faculty of Science \\
  Radboud University, Nijmegen, the Netherlands
}
\publisherline{Springer Nature Switzerland 2021}
\idline{M.~de Bruijne et al.~(Eds.): MICCAI 2021, LNCS 12903}
\publisherdoi{10.1007/978-3-030-87199-4_10}
\ppline{104--113}
\setcounter{page}{104}
\renewcommand{\year}{2021}
\maketitle
\thispagestyle{electronic}
\begin{abstract}
  Multi-view medical image analysis often depends on the combination of information from multiple views.
  However, differences in perspective or other forms of misalignment can make it difficult to combine views effectively, as registration is not always possible.
  Without registration, views can only be combined at a global feature level, by joining feature vectors after global pooling.
  We present a novel cross-view transformer method to transfer information between unregistered views at the level of spatial feature maps.
  We demonstrate this method on multi-view mammography and chest X-ray datasets.
  On both datasets, we find that a cross-view transformer that links spatial feature maps can outperform a baseline model that joins feature vectors after global pooling.

\end{abstract}
\section{Introduction}

Many medical imaging tasks use data from multiple views or modalities, but it can be difficult to combine those effectively.
While multi-modal images can usually be registered and treated as multiple input channels in a neural network, images from different views can be difficult to register correctly (e.g., \cite{Carneiro2017}).
Therefore, most multi-view models process views separately and only combine them after global pooling, which removes any local correlations between views.
If these local correlations are important for the interpretation of the images, models could be improved by linking views at an earlier, spatial feature level.

We discuss two example tasks: mammography classification with craniocaudal (CC) and mediolateral oblique (MLO) views, and chest X-ray classification with frontal and lateral views.
In both applications, multi-view models can outperform single-view models (e.g., \cite{Hashir2020,Wu2019})
However, the different perspectives make registration challenging and make a channel-based approach unsuitable.

We propose a method that can link unregistered views at the level of spatial feature maps.
Inspired by the attention-based transformer models \cite{Vaswani2017} that model links between distant parts of a sequence or image, our model uses attention to link relevant areas between views.
We apply this transformer to the intermediate feature maps produced by a CNN.
Based on a trainable attention mechanism, the model retrieves features from one view and transfers them to the other, where they can be used to add additional context to the original view.

Our approach does not require pixel-wise correspondences -- it compares all pixels in the feature maps from view A to all pixels in the feature maps from view B -- but combines views using a trainable attention model.
By applying this to feature maps instead of directly to the input, we allow the model to link higher-level features and reduce computational complexity.
Since linking all pixel pairs can still be expensive, we additionally investigate an alternative implementation that groups pixels with similar features in visual tokens \cite{Wu2020}.

In this paper, we present these novel pixel-wise and token-based cross-view transformer approaches and apply them to two public datasets.
Although combining features after global pooling is a relatively common way to handle multi-view information with unregistered medical images, to our knowledge there are no methods that use a transformer-based approach to do this at the spatial feature level.
The proposed model can be easily embedded as a module within baseline multi-view architectures that combine views after global pooling.
We evaluate our method on the CBIS-DDSM mammography dataset \cite{Heath2001,Lee2017} and the CheXpert chest X-ray dataset \cite{Irvin2019}.
Based on our experiments, we think that early combination of features can improve the classification of multi-view images.

\section{Related Work}

There are many works on multi-view classification of medical images.
In this section we focus on methods applied to mammography and chest X-ray data.
Most methods combine views at a late stage, usually by concatenating feature vectors obtained from the different views, followed by a fully connected part to make a final prediction.
We use this approach in our multi-view baseline.

Combining features at a global level is common for mammography images, which are difficult to register \cite{Carneiro2017}.
For example, Bekker et al.~\cite{Bekker2015} combined binary predictions from view-specific classifiers.
Carneiro et al.~\cite{Carneiro2017} combined feature from view-specific CNN branches after global pooling.
Wu et al.~\cite{Wu2019} discuss multiple ways to combine views in a single network, all with view-specific convolution branches.
Similar architectures were proposed elsewhere (e.g., \cite{NasirKhan2019,Sun2019}).

Other works combine views at a regional level.
Wang et al.~\cite{Wang2018b} proposed a region-based three-step method: after extracting mass ROIs (regions of interest) from each view, they used a CNN with an attention-driven approach to extract view-specific features from each ROI.
Finally, the features from both views are combined with additional clinical features by an LSTM-based fusion model.
Similarly, Ma et al.~\cite{Ma2019} proposed using Faster RCNNs to detect ROIs in each view, which they then converted to feature vectors and combined in a multi-view network.
Liu et al.~\cite{Liu2020} used a model with bipartite graph convolution to link views based on pseudo landmarks, while satisfying geometric constraints.

Most similar to our approach is the work by Zhao et al.~\cite{Zhao2020}, who applied a joint attention mechanism that combined two views or two sides (left and right breasts) to produce channel-wise and spatial attention maps that highlight asymmetric regions.
The outputs of the attention-weighted, view-specific branches are pooled and concatenated to produce a final classification.
Different from our approach, which transfers feature values between views, Zhao et al.~use cross-view information only to compute cross-view attention weights.

For chest X-rays, many datasets only include the frontal view, since the lateral view is harder to read, is mostly useful for specific diagnoses, and is sometimes replaced by a CT scan \cite{Hashir2020}.
Rubin et al.~\cite{Rubin2018} evaluated a model with view-specific convolution branches, global average pooling and a shared fully connected layer, and report that combining frontal and lateral views improved classification performance.
Recently, Hashir et al.~\cite{Hashir2020} compared several multi-view models on a large chest X-ray dataset, showing that while multi-view data is useful for some diagnosis tasks, the frontal view can be sufficient for others.

\section{Methods}

\begin{figure}[t]
  \subfloat[\vphantom{Sj}Single view]{%
    \label{fig:network-diagram:single-view}%
    \includegraphics[height=5cm]{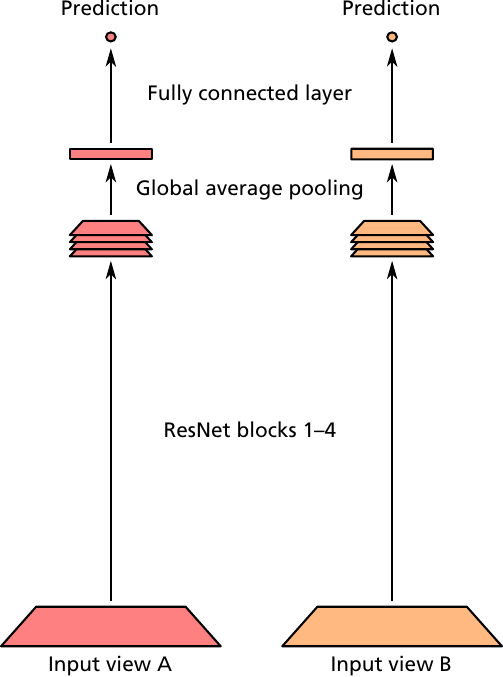}%
  }\hfill%
  \subfloat[\vphantom{Sj}Late join]{%
    \label{fig:network-diagram:late-join}%
    \includegraphics[height=5cm]{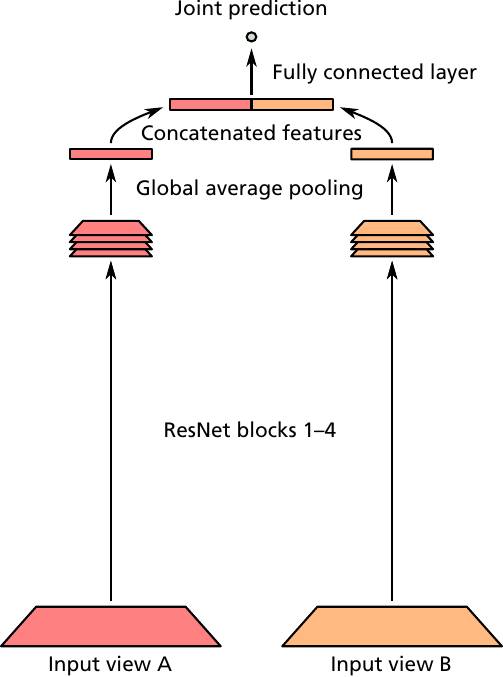}%
  }\hfill%
  \subfloat[\vphantom{Sj}Cross-view transformer]{%
    \label{fig:network-diagram:transformer}%
    \includegraphics[height=5cm]{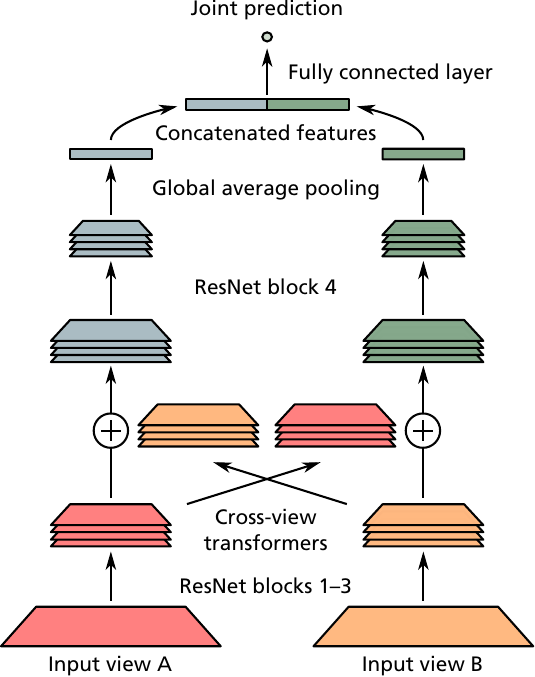}%
  }%
  \caption{Schematic overview of the three architectures.}
  \label{fig:network-diagram}
\end{figure}

In this section, we describe two baseline models and our cross-view transformer models.
All models are designed for a classification task with unregistered, dual-view image pairs.
We use a ResNet-based \cite{He2016} architecture for the view-specific convolution branches, similar to what is used in related work (e.g., \cite{Irvin2019,Wu2019}).

\subsection{Baseline Models}

Our \textit{single-view} baseline (Fig.~\ref{fig:network-diagram:single-view}) follows the basic ResNet architecture.
The network consists of a stack of ResNet blocks with convolution and pooling layers, followed by global average pooling and a fully connected part that computes the final output.
Our \textit{late-join} baseline (Fig.~\ref{fig:network-diagram:late-join}) extends this model to multiple unregistered views, by using a separate convolution branch for each view.
After global pooling, the feature vectors for all views are concatenated and fed into a shared fully connected part to compute the prediction.
This is similar to how multi-view data is commonly combined in other work, such as \cite{Wu2019}.

Depending on the type of data, the view-specific branches can be linked through weight sharing.
This can be a useful regularization if the same low-level features can be used for both views, but might be too restrictive if the views have very different appearances.
We do not use this for our models.

\subsection{Cross-View Transformer Models}

Using view-specific branches and combining feature vectors after global pooling works for unregistered views, but limits the model to learning global correlations.
We propose a transformer-based method that links views at feature map level, without requiring pixel-wise correspondences.
Instead of the self-attention used in standard transformers \cite{Vaswani2017} to transfer information within a single sequence, we use cross-view attention to transfer information between views.
This approach can be used in any dual-view model, such as our late-join baseline.

The \textit{cross-view transformer} (Fig.~\ref{fig:network-diagram:transformer}) works on the intermediate level of the convolutional part of the model.
In our case, we apply the module to the feature maps after the third ResNet block, leaving one ResNet block before global pooling.
At this level, we use the cross-view transformer to transfer features from the source view to the target view.
We make the model bidirectional by applying a second cross-view transformer module in the opposite direction.

We will define two variants of this model: a pixel-wise variant that links pixels from the source and target views, and a token-based variant in which the pixels in the target view are linked to visual tokens in the source view.
For conciseness, we will use the term `pixels' to refer to pixels in the intermediate feature maps.

\subsubsection{Cross-View Attention.}

We use a multi-head attention model \cite{Vaswani2017} with scaled dot-product attention.
For each attention head, we use a $1 \times 1$ convolution with view-specific weights to compute an embedding for the source and target pixels.
We reshape the embedded feature maps for the target view to a query matrix $\mathbf{Q} \in \mathbb{R}^{n \times d}$ and the reshape feature maps for the source view to a key matrix $\mathbf{K} \in \mathbb{R}^{m \times d}$, where $d$ is the size of the embedding and $m$ and $n$ are the number of source and target pixels.
We also reshape the original source feature maps to the value matrix $\mathbf{V} \in \mathbb{R}^{m \times f}$, where $f$ is the number of feature maps.
Next, we use the scaled dot-product attention function \cite{Vaswani2017} to compute
\begin{align}
  \operatorname{Attention} (Q, K, V) = \operatorname{softmax}\left(\frac{\mathbf Q \mathbf K^\top }{\sqrt{d}}\right) \mathbf V \in \mathbb{R}^{n \times f}.
\end{align}
For each target pixel, this computes a weighted sum of the features from the source view, resulting in $f$ new features per attention head.
We reshape the output to $m \times f$ feature maps with the shape of the target feature maps and apply $1 \times 1$ convolution to reduce these to $f$ attention-based feature maps.

We combine the attention-based feature maps $a$ with the original feature maps $x$ of the target data to obtain the combined feature maps $y$:
\begin{align}
  y = \operatorname{LayerNorm} ( x + \operatorname{Dropout} ( \operatorname{Linear} ( a ) ) ),
\end{align}
where $\operatorname{LayerNorm}$ is layer normalization, $\operatorname{Dropout}$ applies dropout, and $\operatorname{Linear}$ a $1 \times 1$ convolution that maps the attention features to the feature space of $x$.
The resulting feature maps $y$ are used as the input for the following ResNet block.

Unlike standard transformer networks \cite{Vaswani2017}, we do not include a positional encoding.
This encoding encodes the relative location of each pixel, allowing the model to distinguish between nearby and faraway pixels within a single image.
This is useful in self-attention for natural images, but is less suitable for cross-view attention.
However, an alternative positional encoding that works across views might help the network to exploit spatial constraints.
We leave this for future work.

\subsubsection{Semantic Visual Tokens.}

Computing the attention between all pairs of source and target pixels can be computationally expensive, even when it is applied to the smaller feature maps at the later stages of the CNN.
We therefore evaluate a variant that uses the tokenization method from \cite{Wu2020} to replace the source pixels with a smaller number of visual tokens, by grouping semantically related pixels.

We apply a three-layer tokenization procedure to the source feature maps.
In the first layer, given flattened feature maps $\mathbf{X} \in \mathbb{R}^{m \times f}$, where $m$ is the number of source pixels and $f$ is the number of feature maps, we compute tokens $\mathbf{T}$:
\begin{align}
  \mathbf{T} = \operatorname{softmax}_{m} \left( \mathbf{X} \mathbf{W}_A \right)^\top \mathbf{X}.
\end{align}
The softmax over the spatial dimension uses the tokenizer weights $\mathbf{W}_A \in \mathbb{R}^{f, L}$ to compute a spatial attention map, which is then used to compute a weighted sum of features for each of the $L$ tokens in $\mathbf{T} \in \mathbb{R}^{L \times f}$.

In the second layer, we use the previous tokens $\mathbf{T}_{in}$ to obtain new tokens $\mathbf{T}$:
\begin{align}
  \mathbf{W}_R = \mathbf{T}_{in} \mathbf{W}_{\mathbf{\mathbf{T} \rightarrow \mathbf{R}}} \\
  \mathbf{T} = \operatorname{softmax}_{m} \left( \mathbf{X} \mathbf{W}_R \right)^\top \mathbf{X}.
\end{align}
where $\mathbf{W}_{\mathbf{\mathbf{T} \rightarrow \mathbf{R}}} \in \mathbb{R}^{f \times f}$.
We repeat this with a different $\mathbf{W}_{\mathbf{\mathbf{T} \rightarrow \mathbf{R}}}$ in the third tokenization layer to obtain the final set of tokens $\mathbf{T}$.
We use these tokens instead of the source pixels in our token-based cross-view transformer.

\section{Data}

\subsubsection{CBIS-DDSM.}

The CBIS-DDSM \cite{Heath2001,Lee2017} is a public mammography dataset with craniocaudal (CC) and mediolateral-oblique (MLO) views with manual annotations.
We solve a binary classification problem on the scans with mass abnormalities, predicting benign vs.~malignant for each CC/MLO pair.
We created five subsets for cross-validation, using stratified sampling while ensuring that all scans for a patient remain in the same subset.
In total, we used image pairs of 708 breasts (636 unique patients), with approximately 46\% labelled malignant.

During preprocessing, we cropped the scans using the method described by Wu et al.~\cite{Wu2019}, using thresholding to position a fixed-size cropping window that includes the breast but excludes most of the empty background.
We downsampled the cropped images to 1/16th of the original resolution to obtain images of $305 \times 188$ pixels.
We normalized the intensities to $\mu=0$ and $\sigma=1$, measured on the nonzero foreground pixels of each scan.

\subsubsection{CheXpert.}

The CheXpert dataset \cite{Irvin2019} is a large public dataset of frontal and lateral chest X-ray scans, annotated for 13 different observations with labels negative, positive, uncertain, or unknown (see supplement).
We used the downsampled version of the dataset as provided on the website.
We selected the visits with complete frontal and lateral views and divided the patients in random subsets for training (23628 samples for 16810 unique patients), validation (3915s, 2802p) and testing (3870s, 2802p).
We normalized the images to $\mu=0$ and $\sigma=1$ and used zero-padding to obtain a constant size of $390 \times 390$ pixels for each view.

To handle the uncertain and unknown labels, we followed \cite{Irvin2019} and used a single network with a three-class softmax output for each task (negative/\hspace{0pt}uncertain/\hspace{0pt}positive).
We excluded the samples with an unknown label from the loss computation for that specific task.
At test time, following \cite{Irvin2019}, we remove the uncertain label and compute the softmax only for the negative and positive outputs.

\section{Experiments}

\subsubsection{Models.}

We compare four models: the single-view model, the late-join model, and the token-based and pixel-based cross-view transformers.
All models use the same ResNet-18 architecture \cite{He2016} for the convolution and pooling blocks up to the global average pooling layer.
We use pre-trained weights on ImageNet, as provided by PyTorch.
After global average pooling, we concatenate the feature vectors for both views and use this as input for a single fully connected layer that computes the output.
(See the supplementary material for a detailed view.)

In the cross-view transformers, we use bidirectional attention and apply the cross-view transformer before the final ResNet block, adding the transformer features to the input for the final convolution and pooling layers.
For the CBIS-DDSM dataset, we evaluated models with 12 or 18 attention heads and 16, 32 or 48 tokens, as well as the pixel-based transformer.
For the CheXpert dataset, we use token-based transformers with 6 or 12 attention heads and 16 or 32 tokens.
In all cases, the embedding size is set to 32 features per head.

\subsubsection{Implementation.}

We implemented the models in PyTorch\footnote{The code for the experiments is available at \\ \url{https://vantulder.net/code/2021/miccai-transformers/}.} and trained using the Adam optimizer with cosine learning rate annealing from 0.0001 to 0.000001, with linear warm-up in the first epochs.
We used rotation, scaling, translation, flipping and elastic deformations as data augmentation.

On the CBIS-DDSM dataset, we used a weighted binary cross-entropy loss to correct for the slight class imbalance.
We trained for 300 epochs (30 warm-up), which was sufficient for all models to converge, and used the model of the final epoch.
We report the mean of three runs over all five folds.

On the CheXpert dataset, we optimized the unweighted cross-entropy loss averaged over all tasks, training for 60 epochs (6 warm-up) to account for the larger dataset.
We computed the AUC-ROC for each task separately and used the performance on the validation set to choose the best epoch for each task.
We report the mean performance over four runs.

We ran all experiments on NVIDIA GeForce RTX 2080 Ti GPUs with 11GB VRAM.
For the CBIS-DDSM dataset, median training times were approximately 23 minutes for a single-view model, 36 minutes for a late-join model, and 37 minutes for the cross-view transformers.
For the much larger CheXpert dataset, we trained for approximately 5 hours per single-view model, versus 10 hours for the late-join and cross-view models.

\begin{table}[t]
  \caption{%
    Area under the ROC curve for the CBIS-DDSM dataset.
    Mean and standard deviation computed over three runs.
    p-values for a two-sided Wilcoxon signed-rank test against the late-join baseline model.
  }
  \vspace{0.5em}
  \label{tab:ddsm-results}
  {
    \setlength{\tabcolsep}{0.5em}
    \begin{tabularx}{\textwidth}{lXcc}
      \toprule
      Model                            &  Views     &  ROC-AUC $\pm$ std.dev. &  p-value  \\
      \midrule
      Single view                      &  CC        &  $0.750 \pm 0.007$      &  0.005    \\
                                       &  MLO       &  $0.763 \pm 0.003$      &  0.036    \\
      Late join                        &  CC + MLO  &  $0.788 \pm 0.008$      &           \\
      Cross-view transformer (tokens)  &  CC + MLO  &  $0.803 \pm 0.007$      &  0.061    \\
      Cross-view transformer (pixels)  &  CC + MLO  &  $0.801 \pm 0.003$      &  0.006    \\
      \bottomrule
    \end{tabularx}
  }

  \vspace{0.5em}

  \caption{%
    Area under the ROC curve for tasks in the CheXpert dataset.
    Mean and standard deviation computed over four runs.
  }
  \label{tab:chexpert-results}
  \vspace{0.5em}
  {
    \setlength{\tabcolsep}{0.5em}
    \begin{tabularx}{\textwidth}{Xcccc}
      \toprule
                                      &  \multicolumn{2}{c}{Single view}              &                         &  Cross-view  \\
      Task                            &  Frontal              &  Lateral              &  Late join              &  (token-based) \\
      \midrule
      Overall                         &  $0.827 \pm 0.007$    &  $0.817 \pm 0.009$    &  $0.829 \pm 0.010$      &  $0.834 \pm 0.002$    \\
      \midrule
      Atelectasis                     &  $0.812 \pm 0.009$    &  $0.809 \pm 0.004$    &  $0.812 \pm 0.016$      &  $0.833 \pm 0.009$    \\
      Cardiomegaly                    &  $0.924 \pm 0.004$    &  $0.902 \pm 0.003$    &  $0.919 \pm 0.003$      &  $0.925 \pm 0.004$    \\
      Consolidation                   &  $0.863 \pm 0.005$    &  $0.848 \pm 0.006$    &  $0.867 \pm 0.004$      &  $0.867 \pm 0.004$    \\
      Edema                           &  $0.882 \pm 0.005$    &  $0.861 \pm 0.005$    &  $0.889 \pm 0.002$      &  $0.889 \pm 0.005$    \\
      Enlarged Cardiomed.             &  $0.812 \pm 0.008$    &  $0.796 \pm 0.003$    &  $0.814 \pm 0.005$      &  $0.810 \pm 0.006$    \\
      Fracture                        &  $0.775 \pm 0.003$    &  $0.764 \pm 0.018$    &  $0.766 \pm 0.019$      &  $0.769 \pm 0.013$    \\
      Lung Lesion                     &  $0.744 \pm 0.013$    &  $0.726 \pm 0.010$    &  $0.747 \pm 0.018$      &  $0.748 \pm 0.007$    \\
      Lung Opacity                    &  $0.808 \pm 0.005$    &  $0.782 \pm 0.006$    &  $0.806 \pm 0.008$      &  $0.805 \pm 0.004$    \\
      Pleural Effusion                &  $0.945 \pm 0.001$    &  $0.946 \pm 0.001$    &  $0.955 \pm 0.002$      &  $0.954 \pm 0.001$    \\
      Pleural Other                   &  $0.789 \pm 0.025$    &  $0.808 \pm 0.030$    &  $0.786 \pm 0.036$      &  $0.803 \pm 0.030$    \\
      Pneumonia                       &  $0.750 \pm 0.004$    &  $0.740 \pm 0.009$    &  $0.766 \pm 0.011$      &  $0.754 \pm 0.008$    \\
      Pneumothorax                    &  $0.869 \pm 0.003$    &  $0.853 \pm 0.004$    &  $0.868 \pm 0.004$      &  $0.872 \pm 0.002$    \\
      Support Devices                 &  $0.773 \pm 0.006$    &  $0.786 \pm 0.015$    &  $0.786 \pm 0.007$      &  $0.803 \pm 0.013$    \\
      \bottomrule
    \end{tabularx}
  }
\end{table}

\section{Results}

On the CBIS-DDSM dataset (Table~\ref{tab:ddsm-results}) the late-join baselines outperformed the single-view baselines.
Adding the cross-view transformer improved the performance further, both for the token-based and pixel-wise variants.
The transformer performance was not very sensitive to the number of heads or tokens: all settings produced similar results (see the table in the supplementary results).

On the CheXpert dataset, the cross-view transformer model also shows an improvement over the baselines (Table~\ref{tab:chexpert-results}).
The improvement is visible for the average performance over all tasks, with the cross-view model performing better than the late-join and single-view frontal models.
The single-view model with the lateral view is less successful.

For individual chest X-ray tasks, the results are quite varied.
For some tasks, such as atelectasis, cardiomegaly and support devices, the cross-view models show an improvement over the late-join model.
For others, such as consolidation and edema, the models are closer together.
This is consistent with observations in other work \cite{Hashir2020}.
In general, with a few exceptions, the cross-view model has a performance that is equal to or better than the late-join models.

\section{Discussion and Conclusion}

The common multi-view approach to merge views after global pooling restricts models to learning global correlations between views.
This may be sufficient for some applications, but a more local approach may be required for others.
This is relatively easy if the images are spatially aligned and can be treated as multiple input channels, but is difficult when there are different perspectives or other misalignments that make it impossible to register the images correctly.

In this paper, we proposed a cross-view transformer approach to link unregistered dual-view images based on feature maps.
Our experiments on two unregistered multi-view datasets indicate that this approach can outperform a model that links views on a global level.
The cross-view transformer module is easy to integrate in any multi-view model with view-specific convolution branches.

Whether a cross-view transformer approach is useful depends on the application, since some tasks will benefit more from multi-view information and local correlations than others.
We can see an indication of this in the results for the chest X-ray data (Table~\ref{tab:chexpert-results}), where the late-join and cross-view transformer models sometimes do and sometimes do not have an advantage over the single-view models.
This is consistent with results from Hashir et al.~\cite{Hashir2020}, who made similar observations about multi-view features on a different chest X-ray dataset.

The cross-view transformer mechanism can be computationally expensive.
This can be reduced by applying the transformer later in the network, when the number of pixels is smaller.
The memory requirements can be reduced with an efficient gradient implementation, by recomputing the pairwise attention scores on the fly.
Using the token-based approach further reduces the requirements.
In practice, we found that the additional computational and memory requirements were relatively limited, compared with those of the convolution layers.

For this paper we focussed on an evaluation of the cross-view transformer, presenting experiments with downsampled images and with the relatively small CBIS-DDSM dataset.
While this allowed us to run more experiments, a better absolute performance could be achieved with higher-resolution data and more training images.
For mammography classification, the state-of-the-art methods \cite{Wu2019} use similar ResNet-based architectures, trained with different and larger datasets.
For chest X-rays, the best-performing methods on the CheXpert dataset \cite{Irvin2019} use the full-resolution dataset and ensemble methods.

In summary, we presented a novel cross-view transformer approach that can transfer information between views, by linking feature maps before global pooling.
On two datasets, we found that combining multi-view information on a spatial level can achieve better results than a model that merges features at an image level.
We believe this can be an interesting addition for models that need to learn inter-view correlations in applications with unregistered images.

\section*{Acknowledgments}

The research leading to these results is part of the project ``MARBLE'', funded from the EFRO/OP-Oost under grant number PROJ-00887.
Some of the experiments were carried out on the Dutch national e-infrastructure with the support of SURF Cooperative.

\bibliographystyle{splncs04}

\begin{thebibliography}{10}
\providecommand{\url}[1]{\texttt{#1}}
\providecommand{\urlprefix}{URL }
\providecommand{\doi}[1]{https://doi.org/#1}

\bibitem{Bekker2015}
Bekker, A.J., Shalhon, M., Greenspan, H., Goldberger, J.: Multi-{{View
  Probabilistic Classification}} of {{Breast Microcalcifications}}. IEEE
  Transactions on Medical Imaging  \textbf{35}(2),  645--653 (Oct 2015).
  \doi{10.1109/TMI.2015.2488019}

\bibitem{Carneiro2017}
Carneiro, G., Nascimento, J., Bradley, A.P.: Deep learning models for
  classifying mammogram exams containing unregistered multi-view images and
  segmentation maps of lesions. In: Zhou, S.K., Greenspan, H., Shen, D. (eds.)
  Deep {{Learning}} for {{Medical Image Analysis}}, pp. 321--339. {Academic
  Press} (Jan 2017). \doi{10.1016/B978-0-12-810408-8.00019-5}

\bibitem{Hashir2020}
Hashir, M., Bertrand, H., Cohen, J.P.: Quantifying the {{Value}} of {{Lateral
  Views}} in {{Deep Learning}} for {{Chest X}}-rays. In: Medical {{Imaging}}
  with {{Deep Learning}}. pp. 288--303. {PMLR} (Sep 2020)

\bibitem{He2016}
He, K., Zhang, X., Ren, S., Sun, J.: Deep {{Residual Learning}} for {{Image
  Recognition}}. In: {{CVPR}}. pp. 770--780 (2016).
  \doi{10.3389/fpsyg.2013.00124}

\bibitem{Heath2001}
Heath, M., Bowyer, K., Kopans, D., Moore, R., Jr, P.K.: The {{Digital
  Database}} for {{Screening Mammography}} (2001)

\bibitem{Irvin2019}
Irvin, J., Rajpurkar, P., Ko, M., Yu, Y., {Ciurea-Ilcus}, S., Chute, C.,
  Marklund, H., Haghgoo, B., Ball, R., Shpanskaya, K., Seekins, J., Mong, D.A.,
  Halabi, S.S., Sandberg, J.K., Jones, R., Larson, D.B., Langlotz, C.P., Patel,
  B.N., Lungren, M.P., Ng, A.Y.: {{CheXpert}}: {{A Large Chest Radiograph
  Dataset}} with {{Uncertainty Labels}} and {{Expert Comparison}}.
  arXiv:1901.07031 [cs, eess]  (Jan 2019)

\bibitem{Lee2017}
Lee, R.S., Gimenez, F., Hoogi, A., Miyake, K.K., Gorovoy, M., Rubin, D.L.: A
  curated mammography data set for use in computer-aided detection and
  diagnosis research. Scientific Data  \textbf{4} (2017).
  \doi{10.1038/sdata.2017.177}

\bibitem{Liu2020}
Liu, Y., Zhang, F., Zhang, Q., Wang, S., Wang, Y., Yu, Y.: Cross-{{View
  Correspondence Reasoning Based}} on {{Bipartite Graph Convolutional Network}}
  for {{Mammogram Mass Detection}}. In: 2020 {{IEEE}}/{{CVF Conference}} on
  {{Computer Vision}} and {{Pattern Recognition}} ({{CVPR}}). pp. 3811--3821.
  {IEEE}, {Seattle, WA, USA} (Jun 2020). \doi{10.1109/CVPR42600.2020.00387}

\bibitem{Ma2019}
Ma, J., Liang, S., Li, X., Li, H., Menze, B.H., Zhang, R., Zheng, W.S.:
  Cross-view {{Relation Networks}} for {{Mammogram Mass Detection}}.
  arXiv:1907.00528 [cs]  (Jun 2019)

\bibitem{NasirKhan2019}
Nasir~Khan, H., Shahid, A.R., Raza, B., Dar, A.H., Alquhayz, H.: Multi-{{View
  Feature Fusion Based Four Views Model}} for {{Mammogram Classification Using
  Convolutional Neural Network}}. IEEE Access  \textbf{7},  165724--165733
  (2019). \doi{10.1109/ACCESS.2019.2953318}

\bibitem{Rubin2018}
Rubin, J., Sanghavi, D., Zhao, C., Lee, K., Qadir, A., {Xu-Wilson}, M.: Large
  {{Scale Automated Reading}} of {{Frontal}} and {{Lateral Chest X}}-{{Rays}}
  using {{Dual Convolutional Neural Networks}}. arXiv:1804.07839 [cs, stat]
  (Apr 2018)

\bibitem{Sun2019}
Sun, L., Wang, J., Hu, Z., Xu, Y., Cui, Z.: Multi-{{View Convolutional Neural
  Networks}} for {{Mammographic Image Classification}}. IEEE Access
  \textbf{7},  126273--126282 (2019). \doi{10.1109/ACCESS.2019.2939167}

\bibitem{Vaswani2017}
Vaswani, A., Shazeer, N., Parmar, N., Uszkoreit, J., Jones, L., Gomez, A.N.,
  Kaiser, {\L}., Polosukhin, I.: Attention is all you need. Advances in Neural
  Information Processing Systems  \textbf{30},  5998--6008 (2017)

\bibitem{Wang2018b}
Wang, H., Feng, J., Zhang, Z., Su, H., Cui, L., He, H., Liu, L.: Breast mass
  classification via deeply integrating the contextual information from
  multi-view data. Pattern Recognition  \textbf{80},  42--52 (Aug 2018).
  \doi{10.1016/j.patcog.2018.02.026}

\bibitem{Wu2020}
Wu, B., Xu, C., Dai, X., Wan, A., Zhang, P., Yan, Z., Tomizuka, M., Gonzalez,
  J., Keutzer, K., Vajda, P.: Visual {{Transformers}}: {{Token}}-based {{Image
  Representation}} and {{Processing}} for {{Computer Vision}}. arXiv:2006.03677
  [cs, eess]  (Nov 2020)

\bibitem{Wu2019}
Wu, N., Phang, J., Park, J., Shen, Y., Huang, Z., Zorin, M., Jastrzebski, S.,
  Fevry, T., Katsnelson, J., Kim, E., Wolfson, S., Parikh, U., Gaddam, S., Lin,
  L.L.Y., Ho, K., Weinstein, J.D., Reig, B., Gao, Y., Pysarenko, H.T.K., Lewin,
  A., Lee, J., Airola, K., Mema, E., Chung, S., Hwang, E., Samreen, N., Kim,
  S.G., Heacock, L., Moy, L., Cho, K., Geras, K.J.: Deep {{Neural Networks
  Improve Radiologists}}' {{Performance}} in {{Breast Cancer Screening}}. IEEE
  Transactions on Medical Imaging  (2019). \doi{10.1109/tmi.2019.2945514}

\bibitem{Zhao2020}
Zhao, X., Yu, L., Wang, X.: Cross-{{View Attention Network}} for {{Breast
  Cancer Screening}} from {{Multi}}-{{View Mammograms}}. In: {{ICASSP}} 2020 -
  2020 {{IEEE International Conference}} on {{Acoustics}}, {{Speech}} and
  {{Signal Processing}} ({{ICASSP}}). pp. 1050--1054 (May 2020).
  \doi{10.1109/ICASSP40776.2020.9054612}

\end{thebibliography}

%

\clearpage
\appendix
\section{Supplementary material}

\begin{table}
  {
    \setlength{\tabcolsep}{0.5em}
    \begin{tabularx}{\textwidth}{Xrrrrrrrr}
      \toprule
      Task                            & \multicolumn{2}{l}{Negative} & \multicolumn{2}{l}{Uncertain}
                                      & \multicolumn{2}{l}{Positive} & \multicolumn{2}{l}{Unknown} \\
      \midrule
      Atelectasis                     &     296 &  (0\,\%)   &    3738 & (11\,\%)   &    3539 & (11\,\%)   &   23840 & (75\,\%)   \\
      Cardiomegaly                    &    3175 & (10\,\%)   &    1332 &  (4\,\%)   &    3472 & (11\,\%)   &   23434 & (74\,\%)   \\
      Consolidation                   &    8259 & (26\,\%)   &    3243 & (10\,\%)   &    1733 &  (5\,\%)   &   18178 & (57\,\%)   \\
      Edema                           &    4678 & (14\,\%)   &    1105 &  (3\,\%)   &    2437 &  (7\,\%)   &   23193 & (73\,\%)   \\
      Enlarged \\
      Cardiomediastinum      &    6170 & (19\,\%)   &    2068 &  (6\,\%)   &    1559 &  (4\,\%)   &   21616 & (68\,\%)   \\
      Fracture                        &     498 &  (1\,\%)   &     137 &  (0\,\%)   &    1540 &  (4\,\%)   &   29238 & (93\,\%)   \\
      Lung Lesion                     &     483 &  (1\,\%)   &     365 &  (1\,\%)   &    2091 &  (6\,\%)   &   28474 & (90\,\%)   \\
      Lung Opacity                    &    1512 &  (4\,\%)   &    1141 &  (3\,\%)   &   10969 & (34\,\%)   &   17791 & (56\,\%)   \\
      Pleural Effusion                &    9850 & (31\,\%)   &    2002 &  (6\,\%)   &    9002 & (28\,\%)   &   10559 & (33\,\%)   \\
      Pleural Other                   &     107 &  (0\,\%)   &     822 &  (2\,\%)   &     996 &  (3\,\%)   &   29488 & (93\,\%)   \\
      Pneumonia                       &     899 &  (2\,\%)   &    2690 &  (8\,\%)   &    1318 &  (4\,\%)   &   26506 & (84\,\%)   \\
      Pneumothorax                    &    8293 & (26\,\%)   &     428 &  (1\,\%)   &    1703 &  (5\,\%)   &   20989 & (66\,\%)   \\
      Support Devices                 &     773 &  (2\,\%)   &     156 &  (0\,\%)   &    8570 & (27\,\%)   &   21914 & (69\,\%)   \\
      \bottomrule
    \end{tabularx}
  }
  \vspace{0.5em}
  \caption{%
    Number of samples per class in the CheXpert dataset.
  }
\end{table}

\begin{table}
  {
    \setlength{\tabcolsep}{0.5em}
    \begin{tabular}{lccccc}
      \toprule
      Layer                  & Input channels   & Output channels  & Kernel size  & Stride        & Padding       \\
      \midrule
      Conv 1                 & 3                & 64               & $7 \times 7$ & $2 \times 2$  & $3 \times 3$  \\
      BatchNorm              & 64               & 64               &              &               &               \\
      ReLU                   & 64               & 64               &              &               &               \\
      MaxPool                & 64               & 64               & $3 \times 3$ & $2 \times 2$  & $1 \times 1$  \\
      ResNet block 1         & 64               & 64               & $3 \times 3$ &               & $1 \times 1$  \\
      ResNet block 2         & 64               & 128              & $3 \times 3$ & $2 \times 2$  & $1 \times 1$  \\
      ResNet block 3         & 128              & 256              & $3 \times 3$ & $2 \times 2$  & $1 \times 1$  \\
      ResNet block 4         & 256              & 512              & $3 \times 3$ & $2 \times 2$  & $1 \times 1$  \\
      Global average pooling & 512              & 512              & $3 \times 3$ & $2 \times 2$  & $1 \times 1$  \\
      Fully connected layer  & 512              & 1                &              &               &               \\
      Sigmoid                & 1                & 1                &              &               &               \\
      \bottomrule
    \end{tabular}
  }
  \vspace{0.5em}
  \caption{%
    Architecture of the single-view network based on ResNet-18.
    The other networks use the same convolutional architecture.
  }
\end{table}

\begin{table}
  {
    \setlength{\tabcolsep}{0.5em}
    \begin{tabularx}{\textwidth}{Xcc}
      \toprule
                           & \multicolumn{2}{c}{Attention heads}               \\
                           &     12                   &     18                  \\
      \midrule
      16 tokens            &   $ 0.793 \pm 0.002 $    &   $ 0.799 \pm 0.018 $    \\
      32 tokens            &   $ 0.796 \pm 0.005 $    &   $ 0.798 \pm 0.002 $    \\
      48 tokens            &   $ 0.802 \pm 0.002 $    &   $ 0.798 \pm 0.007 $    \\
      Pixels               &   $ 0.799 \pm 0.007 $    &   $ 0.804 \pm 0.004 $    \\
      \bottomrule
    \end{tabularx}
  }
  \vspace{0.5em}
  \caption{%
    Area under the ROC curve for the CBIS-DDSM dataset for different configurations of the token-based and pixel-wise cross-view transformers.
    Mean and standard deviation computed over three runs.
  }
\end{table}

\end{document}